\documentclass[runningheads]{llncs}
\usepackage{graphicx}
\usepackage{amsmath} 

\DeclareMathOperator*{\argmin}{arg\,min}

\usepackage{xcolor} 
\usepackage{soul}
\usepackage[separate-uncertainty=true]{siunitx} 

\usepackage{hyperref} 

\usepackage[draft]{todonotes}

\begin{document}
    \title{Intraoperative Liver Surface Completion with Graph Convolutional VAE}
    %
    %
    \author{
        Simone Foti \inst{1,2} \and
        Bongjin Koo \inst{1,2} \and
        Thomas Dowrick \inst{1,2} \and
        Jo\~ao Ramalhinho \inst{1,2} \and
        Moustafa Allam \inst{3} \and 
        Brian Davidson \inst{3} \and
        Danail Stoyanov \inst{1,2} \and \\
        Matthew J. Clarkson \inst{1,2}
    }
    \authorrunning{Foti et al.}

    \institute{
        Wellcome/EPSRC Centre for Interventional and Surgical Sciences, \\ University College London, London, UK \and
        Centre For Medical Image Computing, University College London, London, UK \and
        Division of Surgery and Interventional Science, University College London, London, UK  \\
        \email{s.foti@cs.ucl.ac.uk}
    }
    
    \maketitle              
    
    \begin{abstract}
        In this work we propose a method based on geometric deep learning to predict the complete surface of the liver, given a partial point cloud of the organ obtained during the surgical laparoscopic procedure. We introduce a new data augmentation technique that randomly perturbs shapes in their frequency domain to compensate the limited size of our dataset. The core of our method is a variational autoencoder (VAE) that is trained to learn a latent space for complete shapes of the liver.
        At inference time, the generative part of the model is embedded in an optimisation procedure where the latent representation is iteratively updated to generate a model that matches the intraoperative partial point cloud. The effect of this optimisation is a progressive non-rigid deformation of the initially generated shape. 
        Our method is qualitatively evaluated on real data and quantitatively evaluated on synthetic data. We compared with a state-of-the-art rigid registration algorithm, that our method outperformed in visible areas.
    
        \keywords{Laparascopic Liver Surgery \and Geometric Deep Learning \\ \and Graph Convolution \and Surface Completion \and Variational Autoencoder.}
    \end{abstract}
    
    \section{Introduction}
        \label{sec:intro}
        The loss of direct vision and tactile feedback in laparoscopic procedures introduces an additional level of complexity for surgeons. Augmented reality (AR) is a promising approach to alleviate these limitations and provide guidance during the procedure. However, it remains an open challenge for laparoscopic surgery of the liver, which is one of the largest and most deformable organs. AR is usually achieved by registering a preoperative 3D model to the intraoperative laparoscopic view. Clinically available state-of-the-art commercial systems use manual point-based rigid registration~\cite{prevost2019efficiency}, while recent research works focus on either rigid~\cite{luo2019augmented,robu2018global} or deformable~\cite{brunet2019physics,espinelcombining,koo2017deformable,ozgur2018preoperative} registration techniques requiring different amounts of manual interactions and computations on the preoperative data. In contrast, we formulate the deformable registration as a shape completion problem that does not rely on patient specific preoperative computations. 
        
        Even though the underlying techniques are different, the common application and presence of an optimisation procedure make our method similar to registration. We believe that our method has the potential to become a precursor to a new approach for registration. Thus, we directly compare our method with a rigid registration algorithm (Go-ICP~\cite{yang2015go}) that aligns two point clouds. This algorithm was successfully used for laparoscopic liver applications in~\cite{luo2019augmented}, where the preoperative model was registered onto the intraoperative point cloud obtained using an unsupervised neural network for depth estimation. Our method is similar, but relying on a manual interaction it is also able to predict a deformed model that better fits the point cloud. Other methods, such as~\cite{koo2017deformable,ozgur2018preoperative} attempt the registration of preoperative models directly on the intraoperative images requiring manual image annotations. Even though they still show high errors in areas not visible from the camera, these methods showed extremely good performances in coping with deformations. Both use biomechanical models to simulate deformations, and~\cite{ozgur2018preoperative} requires an additional preoperative step where multiple possible patient-specifc simulations have to be performed. 
        
        The most similar works to ours are~\cite{abdi2019variational} and~\cite{litany2018deformable}. The former leverages a voxel-based conditional variational autoencoder (VAE) to complete missing segments of bone and plan jaw reconstructive surgical procedures. Not only the anatomical structures considered in their work are not deformable and the missing segments are small compared to the complete shape, but also their solution is constrained by the remaining healthy structures. On the other hand, our problem is more ill-posed because the liver is highly deformable and the missing parts are much larger than the partial intraoperative shape. In addition, voxel-based representations of shapes are inefficient volume representations that struggle to achieve high resolutions and to handle deformations. Therefore, we represent 3D shapes as Riemannian manifolds discretised into meshes and use geometric deep learning techniques to process these data. 
        In particular, our work adapts~\cite{litany2018deformable} to achieve shape completion in laparoscopic liver surgery by (\emph{i}) overcoming the shortage of data; (\emph{ii}) compensating the lack of point correspondences between partial and complete shapes; (\emph{iii}) redefining the VAE training loss to deal with non-uniformly sampled meshes; and (\emph{iv}) leveraging preoperative data for the initialisation.
        %
        The optimisation process for shape completion, makes the methodology suitable for registration, but there are a few key steps that need innovating.
        We believe this is the first attempt to bring geometric deep learning in to computer assisted interventions.
        
    \section{Methods}
        \label{sec:methods}
\begin{figure}
            \centering
            \includegraphics[width=\textwidth]{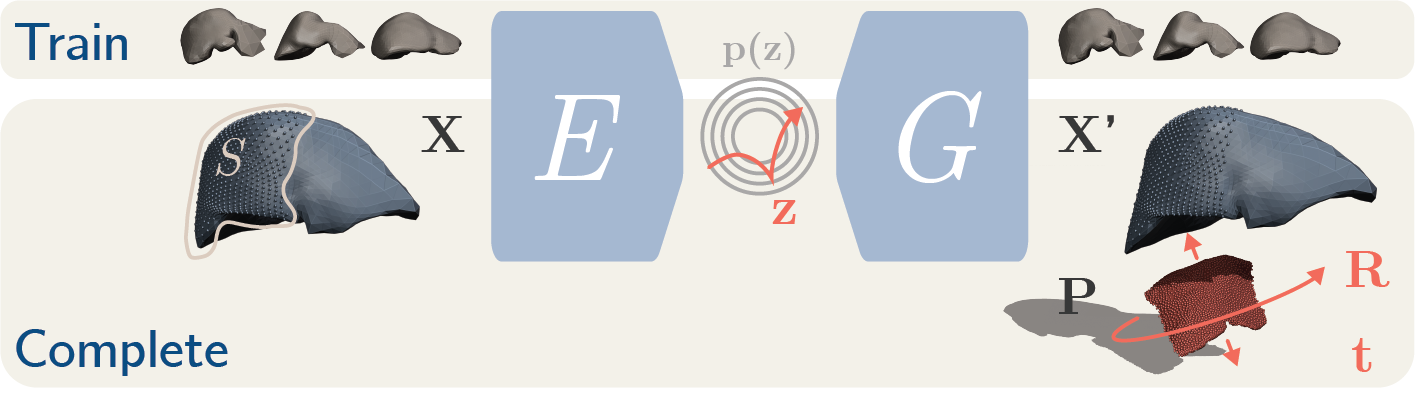}
            \caption{\textbf{Schematic Description of the Proposed Method.} \textit{Top:} a VAE ($\{E, G\}$) is trained on complete preoperative meshes of the liver. \textit{Bottom:} the shape completion starts with a manual selection $S$ on the preoperative mesh $\mathbf{X}$. The latent representation obtained encoding  $\mathbf{X}$ is used to initialise $\mathbf{z}$. The error between the selection on the generated mesh ($\mathbf{X}'_S=S \circ G(\mathbf{z})$) and the partial intraoperative point cloud $\mathbf{P}$ is minimised optimising over $\mathbf{z}, \mathbf{R}, \mathbf{t}$, which control the shape of new generated meshes, the rotation of $\mathbf{P}$ and its translation respectively.}
            \label{fig:method}
        \end{figure}

        The proposed method (Fig.~\ref{fig:method}) estimates the complete mesh of a liver given a partial point cloud of its surface. A graph convolutional variational autoencoder is trained to generate complete shapes (Sec.~\ref{sec:shape_gen}) and a distinct optimisation procedure non-rigidly deforms them to fit the partial point cloud (Sec.~\ref{sec:shape_compl}). 
        
        \subsection{Shape Generator}
            \label{sec:shape_gen}
            A 3D mesh can be represented as a graph $\mathcal{M} = \{\mathbf{X}, \boldsymbol{\varepsilon}\}$, where $\mathbf{X} \in \bbbr^{N \times 3}$ is its vertex embedding and $\boldsymbol\varepsilon \in \bbbn^{\epsilon \times 2}$ is the edge connectivity that defines its topology. Traditional convolutional operators, well suited for grid data such as images and voxelizations, are generally incompatible with the non-Euclidean domain of graphs. Following~\cite{litany2018deformable}, we chose to build our generative model with the Feature-Steered graph convolutions defined in~\cite{verma2018feastnet}. This operator dynamically assigns filter weights to k-ring neighbourhoods according to the features learned by the network. In particular, given a generic feature vector field where for each vertex $i$ we have a feature vector $\mathbf{x}_i$, we can define the output of the convolutional operator as
            
            \begin{equation}
                \mathbf{y}_i = \mathbf{b} + \frac{1}{|\mathcal{N}_i|}\sum_{j \in \mathcal{N}_i} \sum_{m=1}^M q_m(\mathbf{x}_i, \mathbf{x}_j) \mathbf{W}_m \mathbf{x}_j
            \end{equation}
            
            \noindent where $ \mathbf{b}$ is a learnable bias, $q_m(\mathbf{x}_i, \mathbf{x}_j)$ is a translation-invariant assignment operator that, using a soft-max over a linear transformation of the local feature vectors, learns to dynamically assign $\mathbf{x}_j$ to the $m$-th learnable weight matrix $\mathbf{W}_m$, and $\mathcal{N}_i$ is the neighbour of the $i$-th vertex with cardinality $|\mathcal{N}_i|$.
            
            Every VAE is made of an encoder-decoder pair, where the decoder is used as a generative model and is usually referred to as generator. Following this convention, we define our architecture as a pair $\{E, G\}$. Let $\mathcal{X}_c$ be the vertex embedding domain of complete shapes and $\mathcal{Z}$ the latent distribution domain. Then, the two networks are defined as two non-linear functions such that $E: \mathcal{X}_c \rightarrow \mathcal{Z}$ and $G: \mathcal{Z} \rightarrow \mathcal{X}_c$. With $\mathbf{X} \in \mathcal{X}_c$ and $\mathbf{z} \sim \mathcal{Z}$, the generator is described by the likelihood $p(\mathbf{X}| \mathbf{z})$ while the encoder is defined as a variational distribution $q(\mathbf{z}|\mathbf{X})$ that approximates the intractable model posterior distribution. Both $E$ and $G$ are parametrised by neural networks whose building blocks are the Feature-Steered graph convolutions. During training, a reconstruction loss ($\mathcal{L}_{recon}$) encourages the output of the VAE to be as close as possible to its input and a regularisation term ($\mathcal{L}_{KL}$) pushes the variational distribution towards the prior distribution $p(\mathbf{z})$, which is defined as a standard spherical Gaussian distribution. While we set $\mathcal{L}_{KL}$ to be the Kullback–Leibler (KL) divergence, we define $\mathcal{L}_{recon}$ as a vertex-density-weighted mean-squared-errors loss. Let $\mathbf{x}_i$ be the i-th position vector (i.e. a feature vector of size 3) and $\mathbf{x}'_i$ its corresponding point in $\mathbf{X}'=G(E(\mathbf{X}))$. We have:
            
            \begin{equation}
                \label{eq:loss}
                    \mathcal{L}_{recon} = \frac{1}{N} \sum_{i=0}^{N-1} \boldsymbol{\gamma}\|\mathbf{x}'_i - \mathbf{x}_i \|_2^2 \quad \text{with} \; \boldsymbol{\gamma} \propto \frac{1}{\mathcal{N}_i} \sum_{j \in \mathcal{N}_i} \|\mathbf{x}_i - \mathbf{x}_j \|_2^2
            \end{equation}
            
            \noindent where $\gamma$ is a vertex-wise weight that increases the contribution of the errors in low vertex-density regions, thus preventing the generated mesh from fitting only densely sampled areas. The total loss is then computed as linear combination of the two terms: $\mathcal{L}_{tot} = \mathcal{L}_{recon} + \alpha \mathcal{L}_{KL}$.

            \subsubsection{Data Preparation}
                Though the chosen graph convolution was proven effective also on datasets with different graph topologies~\cite{verma2018feastnet}, we decided to remesh our data in order to have the same topology and known point correspondences across all the preoperative meshes and all the generated shapes. This accelerates and eases the training procedure, allowing us to define a simple and computationally-efficient loss function (Eq.~\ref{eq:loss}). In addition, thanks to the consistent vertex indexing it is possible to easily perform the initial manual selection described in Sec.~\ref{sec:shape_compl}. 
                
                In order to consistently remesh our dataset, we run an optimisation procedure that iteratively deforms an ico-sphere with a predefined topology and fixed number of vertices. Following~\cite{wang2018pixel2mesh}, the loss function is given by $\mathcal{L}_{remesh} = \lambda_0 \mathcal{L}_{Ch} + \lambda_1 \mathcal{L}_{n} + \lambda_2 \mathcal{L}_{L} + \lambda_3 \mathcal{L}_{E}$. $\mathcal{L}_{Ch}$ is the Chamfer distance that averages the distances between each point in a mesh and the closest point in the other mesh, and vice versa. $\mathcal{L}_{n}$ is the normal loss that requires the edge between a vertex and its neighbours to be perpendicular to the normal of the closest vertex in the ground truth. $\mathcal{L}_{L}$ is the mesh Laplacian regularisation loss that avoids self-intersections, acting as a surface smoothness term. $\mathcal{L}_{e}$ is an edge regularisation that reduces flying vertices by penalising long edges. $\lambda_{0,1,2,3}$ are the weights of each loss term.

            \subsubsection{Spectral Augmentation}
                The small size of our dataset makes it difficult to train a generative model that can generalise to new shapes. Simple shape augmentation techniques such as random rotations, translations and scalings can be used to augment the dataset, but shapes are not deformed and the performance gain is therefore limited. Instead of attempting an augmentation in the spatial domain we propose a data augmentation technique that operates in the frequency domain, 
                which is a known concept in the literature \cite{rong2008spectral}. However, 
                we simplify and randomise the spectral deformation making it suitable for data augmentation.
                We thus compute the un-normalized graph Laplacian operator $\mathbf{L} = \mathbf{D} - \mathbf{A}$, where $\mathbf{A} \in \bbbr^{N \times N}$ is the adjacency matrix and $\mathbf{D} \in \bbbr^{N \times N}$ is the diagonal degree matrix with $D_{ii} = \sum_j A_{ij}$. Computing the eigenvalue decomposition of the Laplacian, $\mathbf{L} = \mathbf{U \Lambda U}^T$, we obtain a set of orthonormal eigenvectors (columns of $\mathbf{U}$) which are the Fourier bases of the mesh, and a series of eigenvalues (diagonal values of $\mathbf{\Lambda}$) that are its frequencies. The Fourier transform of the vertices can be computed as $\hat{\mathbf{X}} = \mathbf{U}^T\mathbf{X}$ and the inverse Fourier transform as $\mathbf{X} = \mathbf{U}\hat{\mathbf{X}}$~\cite{defferrard2016convolutional}. 
                
                Using these operators, each mesh is transformed into its spectral domain, perturbed, and transformed back to the spatial domain. Hence, the spectral augmented mesh $\mathbf{X}^\dagger$ is computed as $\mathbf{X}^\dagger = \mathbf{U} \boldsymbol{\xi} \mathbf{U}^T\mathbf{X}$, where $\boldsymbol{\xi}$ is a vector that randomly perturbs four mesh frequencies. In particular, the first frequency is never modified because, playing the role of a direct current component \cite{bronstein2017geometric}, it would not deform the shape. One of the following three frequencies, responsible for low frequency variations similar to scalings along the three major axes of the mesh, is arbitrarily perturbed. The remaining three perturbations are applied to randomly selected higher frequencies with the effect of affecting the fine details of the shape.  
                
                It is worth noting that the remeshed data share the same topology, thus the set of orthonormal eigenvectors used to compute the direct and inverse Fourier transforms can be computed one time and then used for every mesh.

        \subsection{Shape Completion}
            \label{sec:shape_compl}
            This section illustrates how a complete shape is obtained from a partial intraoperative point cloud $\mathbf{P} \in \bbbr^{P \times 3}$. 
            In contrast to~\cite{litany2018deformable}, we do not have known (or easily computable) point correspondences between intraoperative point clouds and the generated meshes. Therefore, we relax this assumption at the expense of the introduction of a manual step in the procedure. In fact, the surgeon is asked to roughly select from the preoperative 3D model $\mathbf{X} \in \mathcal{X}_c$ a region of interest that corresponds to the visible surface in the intraoperative image. To reduce computational time and increase robustness against the errors in manual region selection and varying vertex density in the region, we sample the selected vertices with an iterative farthest point sampling~\cite{qi2017pointnet++}, obtaining a selection operator $S$ that gives sparser and uniformly sampled vertices. Since mesh topology consistency is guaranteed by construction, the selected vertices will always have the same indexing for each mesh $\mathbf{X}' \in \mathcal{X}_c$ generated with the model discussed in Sec.~\ref{sec:shape_gen}. The shape completion problem is thus formulated as finding the best latent variable $\mathbf{z}^*$ that generates a complete shape $\mathbf{X}'^*$ plausibly fitting $\mathbf{P}$. Given  $\mathbf{X}'_S = S \circ \mathbf{X}' = S \circ G(\mathbf{z}) \subset \mathbf{X}'$ the subset of selected and sampled vertices from a generated shape, we optimise
            
            \begin{equation}
                \label{eq:opt_completion}
                \min_\mathbf{z, R, t} \mathcal{L}_{Ch}\Big(S \circ G(\mathbf{z}), \; \mathbf{RP} + \mathbf{t} \Big).
            \end{equation}
            
            \noindent It is worth mentioning that not having point correspondences between $\mathbf{P}$ and $\mathbf{X}'_S$ we cannot compute the rotation $\mathbf{R}$ and translation $\mathbf{t}$ in a closed form solution as in~\cite{litany2018deformable}. Therefore, they are iteratively updated alongside $\mathbf{z}$ in the same optimisation procedure. The gradient of the loss in Eq.~\ref{eq:opt_completion} directly influences $\mathbf{R}$ and $\mathbf{t}$, but it needs to be back-propagated through the generator network $G$, without updating the network's weights, to update $\mathbf{z}$. The completion procedure is initialised by centering $\mathbf{P}$ and $\mathbf{X}'_S$, and by setting an initial $\mathbf{z} = \mathbf{z}_0 = E(\mathbf{X})$, thus using as prior the latent representation of the preoperative mesh. The initialisation $\mathbf{z}_0$ can be further refined to $\mathbf{z}^*_0$ by running a few iterations of a second optimisation $\mathbf{z}^*_0 \leftarrow \argmin_\mathbf{z_0} \big(\max_i \| \mathbf{x}_i - \mathbf{x}'_i \|_2^2 \big)$. Finally, by adding to the latent initialisation a small Gaussian noise $\boldsymbol{\eta} \sim \mathcal{G}(\mathbf{0, \Sigma})$ with $\Sigma_{ii} \ll I_{ii}$,  we can generate multiple complete shapes conditioned on the preoperative data and that plausibly fit the intraoperative point cloud $\mathbf{P}$.

    \section{Results}
        \label{sec:results}
        Our dataset consists of 50 meshes of livers which were segmented and reconstructed from preoperative CT scans of different patients. The segmentation and initial mesh generation was performed by \href{https://www.visiblepatient.com/en/}{Visible Patient}. 45 meshes were used to train the VAE, while the remaining 5 meshes were used as a test set to evaluate the network, data augmentation, and shape completion. Given the limited size of the dataset, to not bias results toward the test set, hyperparameters were tuned on the training set. The study was approved by the local research ethics committee (Ref:  14/LO/1264) 
        and written consent obtained from each patient.
        
        The remeshing was performed by deforming an ico-sphere with $2564$ vertices. For this, and all the other optimisations described in this paper, we used the ADAM optimiser~\cite{kingma2014adam}. We remeshed every model with $500$ iterations at a fixed learning rate of $lr = 5e^{-3}$. The weights of the loss function $\mathcal{L}_{remesh}$ were $\lambda_0 = 5$, $\lambda_1 = 0.2 $, $\lambda_2 = 0.3$, and $\lambda_3 = 15$.  
        
        The VAE was built using $M=8$ weight matrices, batch size 20 and latent size 128. LeakyReLU and batch normalisation were used after every layer. The network was implemented in PyTorch Geometric~\cite{Fey/Lenssen/2019} and trained for 200 epochs with $lr=1e^{-3}$ and a KL divergence weight $\alpha = 1e^{-6}$. The training was performed on an NVIDIA Quadro P5000 and took approximately 9 hours.
        
\begin{figure}[t]
            \centering
            \includegraphics[width=\textwidth]{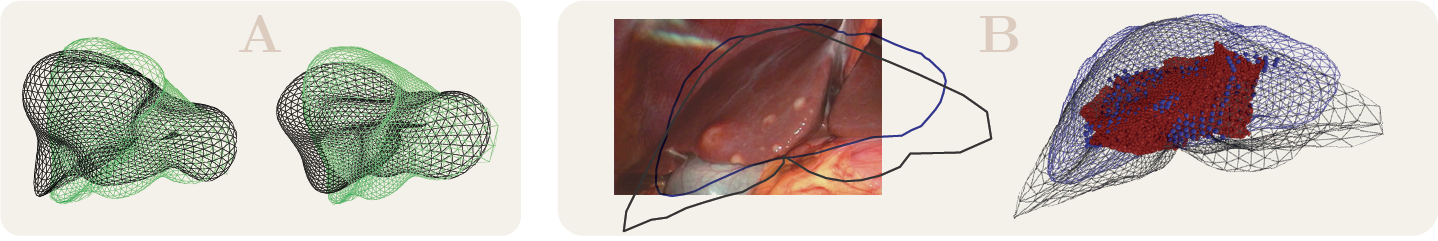}
            \caption{\textbf{Augmentation and Qualitative Results} \textit{A}: effects of the spectral augmentation where a real liver (green) is subject to two different random augmentation (black). \textit{B}: laparoscopic image and comparison between the proposed shape completion (blue) and the Go-ICP registration (black). The intraoperative point cloud is shown in red and the selected point in blue. The contours of the silhouettes are overlaied also on the laparoscopic image.}
            \label{fig:aug_real}
        \end{figure}

        We evaluated the reconstruction performance of the VAE with and without data augmentation while fixing the number of iterations. Applying the spectral augmentation (Fig.~\ref{fig:aug_real}A) we generated 100 new meshes for each model in the training set, thus obtaining 4500 models. An additional online data augmentation composed of random rotations, scalings, and translations was applied. We obtain a mean-squared testing error of $\SI{0.28 \pm 0.04}{\mm}$ when both augmentations are applied, of $\SI{0.45 \pm 0.18}{\mm}$ with the online augmentation alone, of $\SI{0.50 \pm 0.03}{\mm}$ with the spectral augmentation only, and of $\SI{0.92 \pm 0.22}{\mm}$ without any augmentation. 
        We then evaluated the computational cost of the spectral augmentation, finding that when it is performed by computing the Fourier operators for each mesh it takes $0.4532 \pm 0.0568$ seconds per mesh, while, when the operators are precomputed (Sec.~\ref{sec:shape_gen}) the computational time is reduced by one order of magnitude to $0.0487 \pm 0.0092$ seconds per mesh. 
        
        Given the lack of intraoperative 3D ground truths for registration in laparoscopic liver surgery, the evaluation of our method on real data is purely qualitative. In a real operative scenario the computation of a dense and reliable point cloud is still a major challenge. To obtain $\mathbf{P}$ from rectified images of a calibrated laparoscope we used an off-the-shelf depth reconstruction network \cite{chang2018pyramid}. Given the predicted depth map and a manual segmentation of the liver, we first compute $\mathbf{P}$ and then estimate the complete shape $\mathbf{X}'^*$ (Fig.~\ref{fig:aug_real}B).
            
        The quantitative assessment of the shape completion is performed on synthetic data. The five meshes in the test set were manually deformed, trying to reproduce intraoperative liver deformations similar to those expected in a laparoscopic procedure and characterised in \cite{heiselman2017characterization}. To obtain intraoperative partial point clouds, the surface of the deformed models was sampled with vertex selections on three regions: entire front surface $\mathbf{P}_F$, left lobe $\mathbf{P}_L$, and right lobe $\mathbf{P}_R$. Each deformed model is considered the intraoperative ground truth $\mathbf{X}_{GT}$ that we want to infer given a partial point cloud. To maintain a higher $\mathbf{P}$ density, $\mathbf{X}_{GT}$ was not remeshed. Eq.~\ref{eq:opt_completion} is optimised for 100 iterations using ADAM with a different learning rate for each term. To encourage the optimisation over $\mathbf{z}$ and thus the generation of more diverse, progressively deformed meshes $\mathbf{X}'$, we set $lr_z = 5e^{-2}$. The learning rates for $\mathbf{R}$ and $\mathbf{t}$ are  empirically set to $lr_R=1e^{-2}$ and $lr_t=5e^{-5}$. 
        Rotations are regressed faster because the two point clouds were initially centred. In case $\mathbf{z}_0$ is further refined to $\mathbf{z}^*_0$, the same learning rate $lr_z = 5e^{-2}$ is used for 20 iterations. In case multiple complete shape proposals are desired, Eq.~\ref{eq:opt_completion} can be generalised to process batches with a refined initialisation perturbed by $\boldsymbol{\eta}$ with $\Sigma_{ii} = \frac{1}{10}$. 
            
\begin{figure}[t]
            \centering
            \includegraphics[width=\textwidth]{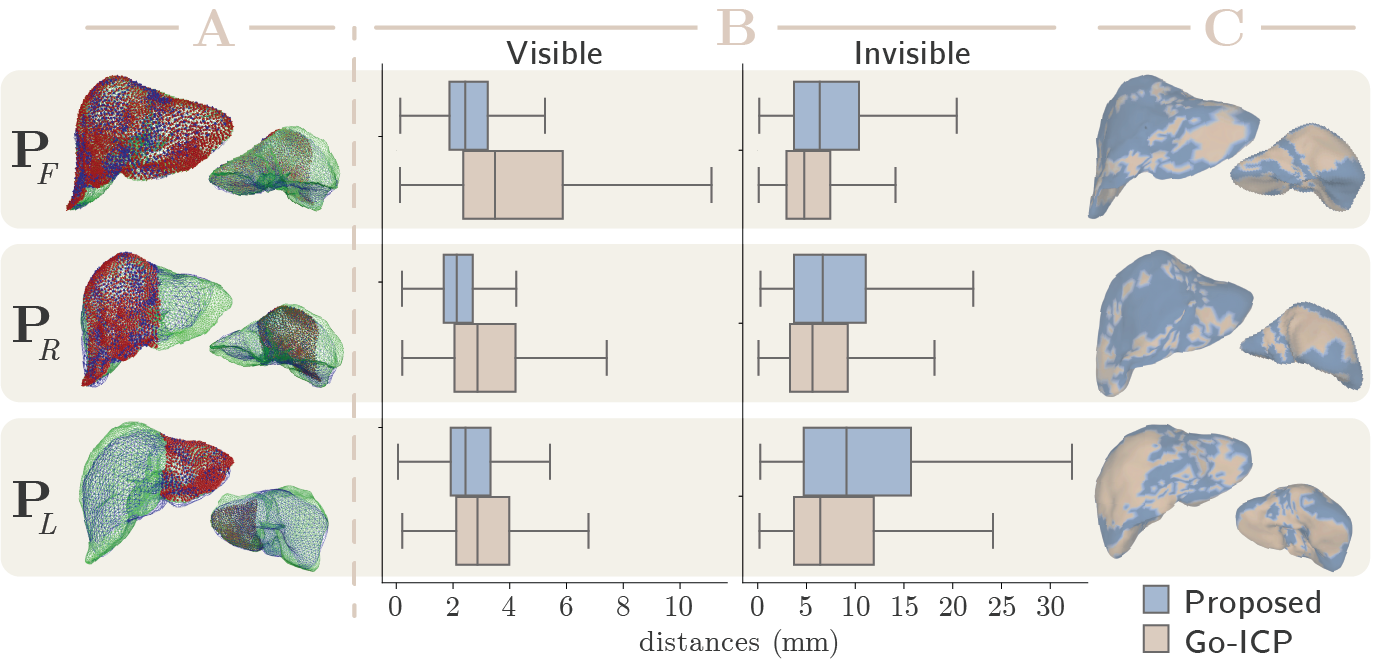}
            \caption{\textbf{Quantitative Results.} \textit{Rows}: results for intraoperative point clouds of front surface $\mathbf{P}_F$, right lobe $\mathbf{P}_R$, and left lobe $\mathbf{P}_L$. \textit{Columns}: A) front and back view of intraoperative point cloud $\mathbf{P}$ (red), intraoperative ground truth $\mathbf{X}_{GT}$ (green), and prediction $\mathbf{X}'^*$ (blue). B) Comparison of vertex-wise distances (\si{mm}) for the proposed method (blue) and Go-ICP (pink). We compute errors separately in the visible and invisible parts of the liver in the camera's field of view. C) Front and back view of $\mathbf{X}'^*$. Colours represent the algorithm with a smaller vertex error.}
            \label{fig:results}
        \end{figure}

        The shape completion was evaluated for each partial shape without $\boldsymbol{\eta}$ perturbation. Since the procedure requires a manual step currently performed with a lasso selection that might affect results, we repeated the evaluation 3 times, for a total of 45 experiments. Selections could be refined and took approximately one minute each.
        We compared our method with the rigid registration using Go-ICP~\cite{luo2019augmented} which has comparable computational time to ours.
        The lack of point correspondences between $\mathbf{X}_{GT}$ and $\mathbf{X}'^*$ does not allow us to evaluate our method using mean-squared errors. Therefore, we define a variation of $\mathcal{L}_{Ch}$ that allows us to compute vertex-wise errors on $\mathbf{X}'^*$.
        For each vertex of one mesh we compute the distance to the closest point on the other mesh. All the distances are assigned to the vertices of $\mathbf{X}'^*$ from which they were computed and are locally averaged.
            Results are reported in Fig.~\ref{fig:results}.
            
    \section{Conclusion}
        \label{sec:conclusion}
        While this work is about shape completion, we believe it could become an alternative to registration or provide a better initialisation for such algorithms.
        From Fig.~\ref{fig:aug_real} we notice that the proposed method seems to fit better the point cloud especially on the left lobe. The lack of a ground truth makes impossible to draw further conclusions from this result. However, observing Fig.~\ref{fig:results} (columns B and C), we can conclude that our method outperforms Go-ICP in visible areas, and, despite performing worse in invisible areas, it predicts a realistic looking model of the liver. Of particular importance is the improvement over visible areas, because these regions are the only ones in the narrow field of interest of the surgeon, where an accurate deformation is required.
        Since the manual selection of the visible area on the preoperative model affects the quality of the results, as future work not only we aim at quantifying the uncertainty involved in the manual interaction, but also at avoiding it by predicting point correspondences between partial and complete shapes. We also believe that the use of biomechanical constraints for deformation could reduce errors in invisible areas.
        In fact, the unconstrained deformations operated by our method through the optimisation of \textbf{z}, despite generating plausible livers fitting the partial intraoperative point cloud, often downscale invisible areas. Even though our method can propose multiple solutions (Sec.~\ref{sec:shape_compl}), identifying the correct complete shape is essential to improve our method and outperform Go-ICP everywhere.
        Thus, we shall also research the introduction of biomechanical constraints while avoiding any patient specific training or simulation. 
        Finally, we are planning to incorporate the liver’s internal structure in our method in order to overlay them on a laparoscopic video during surgery.
        
        \subsubsection{Acknowledgments} 
            This work was supported by the the Wellcome Trust/EPSRC [203145Z/16/Z], and Wellcome Trust / Department of Health [HICF-T4-317]
            

    \bibliographystyle{splncs04}
    \bibliography{bibliography.bib}
\end{document}